\documentclass[sigconf]{acmart}
\usepackage[utf8]{inputenc} 
\usepackage[T1]{fontenc}    
\usepackage{hyperref}       
\usepackage{url}            
\usepackage{booktabs}       
\usepackage{amsfonts}       
\usepackage{nicefrac}       
\usepackage{microtype}      
\usepackage{xcolor}         
\usepackage{algorithm}
\usepackage{booktabs}
\usepackage{multirow}
\usepackage{graphicx}
\usepackage{algorithmic}
\usepackage{subcaption}
\usepackage{balance}

\usepackage{amsthm}
\usepackage{thmtools}
\usepackage{enumitem}

\AtBeginDocument{%
}

\setcopyright{none}
\copyrightyear{2026}
\acmYear{2026}
\acmConference[CIKM '26]{The 35th ACM International Conference on Information and Knowledge Management}{November 2026}{TBD}
\acmBooktitle{The 35th ACM International Conference on Information and Knowledge Management (CIKM '26), November 2026, TBD}
\acmDOI{}
\acmISBN{}

\title{Stability Analysis of Signed Graphs under Structural Perturbations}

\author{Ning Zhang}
\affiliation{%
\institution{University of Oxford}\department{Department of Statistics}
\city{Oxoford}
\country{United Kingdom}}
\email{ning.zhang@stats.ox.ac.uk}

\author{Li Zhang}\authornote{Equal contribution. Author order was determined randomly. Corresponding to: ucesl07@ucl.ac.uk}

\affiliation{%
\institution{University College London}
\institution{Institute of Financial Technology }\department{Department of Civil, Environmental and Geomatic Engineering}
\city{London}
\country{United Kingdom}
}
\email{ucesl07@ucl.ac.uk}

\begin{abstract}
Signed graph learning has emerged as a powerful framework for modeling networks with both positive and negative interactions. While graph-filter-based methods have achieved promising empirical performance on signed networks, their stability under edge perturbations remains less well understood. Existing stability analyses primarily focus on unsigned graphs and do not account for sign-flip perturbations.  
In this paper, we study the stability of signed graph filters under edge additions, deletions, and
sign flips. 
Our theoretical analysis reveals how signed edge perturbations and node feature correlations jointly determine the learned representation stability.
Based on this analysis, we propose a theory-guided perturbation strategy that 
selects structurally influential signed-edge modifications. Experiments on
synthetic and real signed graphs show that the proposed method consistently
induces larger embedding drift, polarity change, and downstream link prediction
degradation than baselines. These
results highlight the importance of \textit{sign-aware} and \textit{distribution-aware} stability analysis for signed
graph learning.

\end{abstract}


\copyrightyear{2026}
\acmYear{2026}
\setcopyright{cc}
\setcctype{by}
\acmConference[CIKM '26]{Proceedings of the 35th ACM International Conference on Information and Knowledge Management}{November 07--11, 2026}{Rome, Italy}
\acmBooktitle{Proceedings of the 35th ACM International Conference on Information and Knowledge Management (CIKM '26), November 07--11, 2026, Rome, Italy}
\acmDOI{10.1145/3799682.3840016}
\acmISBN{979-8-4007-2539-5/2026/11}

\begin{document}

\maketitle

\section{Introduction}


Signed graphs provide a natural representation for networks containing both positive and negative interactions~\cite{leskovec2010signed,derr2018signed}.
In social networks, for example, edge signs can encode the semantics of information propagation: positive edges promote agreement or similarity opinions, whereas negative edges capture conflict. The interplay between these opposing relations gives rise to characteristic structural patterns such as stable patterns of friendship and hostility, as formalized by structural balance theory~\cite{cartwright1956structural,harary1953notion}. Motivated by such relational structure, growing attention has been devoted to learning over signed networks~\cite{pao2026learning}.
A important class of approaches relies on graph filtering, where node representations are obtained by propagating and transforming node features according to signed graph structure.
Despite their empirical success, however, the stability of signed graph filters under structural perturbations remains less well understood.


Existing study on the stability of graph filters and graph neural networks mainly focus on unsigned graphs, where perturbations involve edge additions, deletions, or rewiring~\cite{kenlay2021stability,gao2021stability,zhang2025stability}. In signed networks, however, edges may also be perturbed through sign flipping, which preserves graph connectivity while fundamentally altering the semantics of information propagation. Such perturbations lie outside the scope of classical analyses for unsigned graphs. Furthermore, edge signs are often correlated with node features; for example, nodes with similar opinions are more likely to be connected by positive edges than by negative ones. Yet existing stability analyses do not account for this interaction between node features and signed edge perturbations. This gap motivates the present work.

In this paper, we study the embedding stability of signed graph filters under signed edge perturbations. We focus on embeddings rather than specific downstream tasks, allowing our analysis to provide a general framework for understanding the stability of learning architectures built on graph filters.
Specifically, we follow the probabilistic stability framework of~\cite{zhang2025stability} and characterize the expected embedding change induced by edge perturbations.
We instantiate our theoretical analysis with two representative signed graph filters, the signed adjacency and signed Laplacian filters. 
The resulting stability characterization is both \textit{sign-aware} and \textit{distribution-aware}: the effect of a perturbation depends not only on the modified edges but also on the second-order statistics of the associated node features. Our analysis also reveals how nearby signed edge changes can reinforce or cancel each other depending on their signs and node feature statistics.

Our theoretical analysis naturally leads to a perturbation strategy that scores edge additions, deletions, and sign flips by jointly considering graph structure and node features correlation. The proposed strategy directly targets perturbations that are most influential for signed graph filtering, serving both as an analysis tool for understanding signed graph instability and as an attack mechanism for evaluating the robustness of signed graph filters.  We evaluate the method on both synthetic and real-world signed graphs. The results show that the proposed perturbations induce larger representation changes and stronger downstream degradation than baselines, validating the effectiveness of the stability analysis.




\section{Related Work}

Signed graphs contain positive and negative edges, whose signs encode relational semantics beyond ordinary connectivity, as described by balance and status theories \cite{harary1953balance,leskovec2010signed}. Existing signed graph learning and spectral methods design sign-aware aggregation rules, signed Laplacians, balance/status-based objectives, and signed graph filters \cite{kunegis2010spectral,derr2018sgcn,huang2019sigat,mercado2019powermean,dittrich2020signedgsp,singh2023freq}. These works motivate signed graph operators, but mainly focus on representation learning, clustering, or spectral design rather than perturbation stability.

Analyzing the stability of graph filters has gained increasing attention because graph-based models are often deployed on noisy or incomplete network data. Existing studies investigate different types of graph perturbations, including deterministic edge perturbations, stochastic edge rewiring, graph transfer, graphon perturbations, and adversarial edge attacks \cite{gama2020stability,gao2021stochastic,levie2019transferability,kenlay2021interpretable,ruiz2020graphon,zhang2020feature,ruiz2023transferability,zugner2018nettack,bojchevski2019certifiable}. Most of these works study unsigned graph filters or GNNs, where perturbations are usually modeled as changes in connectivity or in the graph shift operator. Recent probabilistic analyses further measure expected output perturbations and integrate signal statistics into stability analysis \cite{zhang2025stability}. Our work is most closely related to this probabilistic viewpoint, but differs in that we study signed graph filters rather than unsigned graph filters or GCNN embeddings.

\section{Analyzing Stability of Signed Graph
Filters}
\label{sec:theory}
\subsection{Problem formulation}
\paragraph{Signed graphs.}
We use $\mathcal{G}=(\mathcal{V},\mathcal{E}^+,\mathcal{E}^-,\mathbf{A})$ to denote an undirected signed graph on node set $\mathcal{V}$, where $\mathcal{E}^+$ and $\mathcal{E}^-$ are the sets of positive edges and negative edges, respectively.
{We denote by} $\mathbf{A}\in \{\pm1, 0\}^{n\times n}$ its adjacency matrix, where $\mathbf{A}_{ij}=1$ if $(i,j)\in\mathcal{E}^+$, $\mathbf{A}_{ij}=-1$ if $(i,j)\in\mathcal{E}^-$, and $\mathbf{A}_{ij}=0$ otherwise. In this work, we study the stability of learning models under edge perturbations to $\mathcal{G}$. Specifically, we denote the perturbed graph by $\mathcal{G}_p$, with adjacency matrix $\mathbf{A}_p$, and define the edge perturbation matrix as $\Delta\mathbf{A}=\mathbf{A}_p-\mathbf{A}$. For unweighted signed graphs, $\Delta\mathbf{A}_{ij}\in\{-2,-1,0,1,2\}$, where $|\Delta\mathbf{A}_{ij}|=1$ represents signed-edge addition or deletion between nodes $i$ and $j$, while $|\Delta\mathbf{A}_{ij}|=2$ denotes sign flipping. 

\paragraph{Representation learning.} 
Graph-based learning models typically use graph filters to construct node representations (embeddings) that effectively combine node-level information from the input features (graph signals) $\mathbf{X}\in\mathbb{R}^{n\times d}$ with graph topology.
We denote a signed graph filter by $g(\mathcal{G})\in\mathbb{R}^{n\times n}$, such that the resulting node representations are given by $g(\mathcal{G})\mathbf{X}$. In this work, we consider two representative signed graph filters: the signed adjacency filter $g(\mathcal{G})=\mathbf{A}$ and the signed Laplacian filter $g(\mathcal{G})=\mathbf{L}=\mathbf{D}-\mathbf{A}$, where $\mathbf{D}$ is the diagonal degree matrix with $\mathbf{D}_{ii}=\sum_j|\mathbf{A}_{ij}|$.
When the underlying graph is perturbed from $\mathcal{G}$ to $\mathcal{G}_p$, the corresponding filter changes from $g(\mathcal{G})$ to $g(\mathcal{G}_p)$. We denote the resulting filter perturbation by $\mathbf{E}_g=g(\mathcal{G}_p)-g(\mathcal{G})$.


\paragraph{Embedding perturbation.}
In this work, we study the effect of edge perturbations on node representations. Since the resulting representations also depend on the input node features, we evaluate the stability of graph filters from an average-case perspective, following the  framework of~\cite{zhang2025stability}.
Specifically, we assume that the input node features $\mathbf{X}$ are drawn from a distribution $\mathcal{D}$ with second-order statistics $\mathbf{K}=\mathbb{E}[\mathbf{X}\mathbf{X}^{\top}]$. The expected representation perturbation is then given by
\begin{equation}
\mathbb{E}_{\mathbf{X}\sim\mathcal{D}}
\left[
\|g(\mathcal{G})\mathbf{X} -g(\mathcal{G}_p)\mathbf{X}\|_F^2
\right] = 
\left\langle
\mathbf{K},
\mathbf{E}_g^\top\mathbf{E}_g
\right\rangle .
\label{eq:signed_filter_energy}
\end{equation}
Equation~\eqref{eq:signed_filter_energy} shows that the expected representation perturbation depends jointly on the filter perturbation $\mathbf{E}_g$ and input feature statistics $\mathbf{K}$: $\mathbf{E}_g^\top\mathbf{E}_g$ captures how the structural perturbation affects the graph filter; $\mathbf{K}$ weights perturbation directions by
input-signal energy and correlation.
The specific role of edge signs depends on the choice of graph filter, which we examine next.

\subsection{ Expected Perturbation of Signed Graph Filters}
\label{subsec:signed_energy}

We interpret the stability measure in Eq~\eqref{eq:signed_filter_energy} by instantiating it with two representative signed graph filters: the signed adjacency filter and the signed Laplacian filter. For each filter, we characterize how different types of signed-edge perturbations, including edge additions, deletions, and sign flips, contribute to the expected representation perturbation.

\subsubsection{Signed Adjacency Filter}
The signed adjacency filter $g(\mathcal{G})=\mathbf{A}$ takes the input signal $\mathbf{X}$ and outputs $\mathbf{A}\mathbf{X}$, aggregating neighboring signals according to their edge signs, with positive and negative edges contributing with the same and opposite signs, respectively~\cite{kunegis2010spectral, derr2018signed}.

We  denote the set of perturbed vertex pairs by $\mathcal{P}$.  For each perturbed pair $\{u,v\}$, we define the signed edge-change coefficient $ \delta_{uv}= (\mathbf{A}_{p})_{uv} - \mathbf{A}_{uv}$, $\delta_{uv}\in\{-2,-1,0,1,2\}$. 
For undirected graphs, the filter perturbation can be decomposed as $ \mathbf{E}_g = \mathbf{A}_p-\mathbf{A} = \sum_{\{u,v\}\in\mathcal{P}} \delta_{uv} \left( \mathbf{e}_u\mathbf{e}_v^\top +\mathbf{e}_v\mathbf{e}_u^\top\right)$, where $\mathbf{e}_u\in\mathbb{R}^n$ denotes the standard basis vector associated with node $u$.
Thus ($\mathbf{E}_g\mathbf{X})_{u,:} =\sum_{\{u,v\}\in\mathcal{P}}
\delta_{uv}\mathbf{X}_{v,:}$.
Expanding the Frobenius norm gives
\begin{equation}
\begin{aligned}
\|\mathbf{E}_{g}\mathbf{X}\|_F^2
= \hspace{-10pt}
\sum_{\{u,v\}\in\mathcal{P}}
\hspace{-6pt}
\delta_{uv}^2
\bigl(
\|\mathbf{X}_{u,:}\|_2^2
+
\|\mathbf{X}_{v,:}\|_2^2
\bigr) 
+
2 \hspace{-20pt}\sum_{\substack{
\{u,v\},\{u,v'\}\subseteq\mathcal{P}\\
v\neq v'
}}
\hspace{-20pt}
\delta_{uv}\delta_{uv'}
\langle
\mathbf{X}_{v,:},
\mathbf{X}_{v',:}
\rangle .
\end{aligned}
\label{eq:proof_adj_expand_revised}
\end{equation}
Taking expectation and $\mathbf{K}=\mathbb{E}[\mathbf{X}\mathbf{X}^{\top}]$, we can get Proposition 1.

\paragraph{Proposition 1.}
\textit{For the signed adjacency filter $g(\mathbf{\mathcal{G}})=\mathbf{A}$, the expected embedding perturbation \eqref{eq:signed_filter_energy} satisfies}
\begin{equation}
\begin{aligned}
\mathbb{E}_{\mathbf{X}\sim\mathcal{D}}\!\left[\|\mathbf{E}_{g}\mathbf{X}\|_F^2\right]
    =
    \hspace{-10pt}
    \sum_{\{u,v\}\in\mathcal{P}}
    \hspace{-6pt}
    \delta_{uv}^2(\mathbf{K}_{uu}+\mathbf{K}_{vv})
    +
    2\hspace{-16pt}
    \sum_{\substack{\{u,v\},\{u,v'\}\in\mathcal{P}\\ v\neq v'}}
    \hspace{-12pt}
    \delta_{uv}\delta_{uv'}\mathbf{K}_{vv'} .
\end{aligned}
\label{eq:prop1}
\end{equation}

In Eq.~\eqref{eq:prop1}, the first term measures the individual contribution of each edge perturbation to the expected representation distortion. Since the edge-change coefficient has the largest magnitude for a sign flip, i.e., $|\delta_{uv}|=2$, a sign flip contributes four times as much as an edge addition or deletion for the same pair of nodes.
The second term of  Eq.~\eqref{eq:prop1} quantifies the additional contribution arising from pairs of perturbed edges that share a common node, where $\{u,v\}$ and $\{u,v'\}$ intersect at $u$. This term shows that whether two intersecting edge perturbations reinforce or attenuate each other's effects is jointly determined by their perturbation directions through $\delta_{uv}\delta_{uv'}$ and the feature correlation $\mathbf{K}_{vv'}$. In particular, for positively correlated node features $\mathbf{K}_{vv'}>0$, perturbations in the same direction ($\delta_{uv}\delta_{uv'}>0$), e.g., both changing toward more positive edge, reinforce each other, whereas perturbations in opposite directions ($\delta_{uv}\delta_{uv'}<0$) attenuate each other. The opposite behavior occurs when $\mathbf{K}_{vv'}<0$.

\subsubsection{Signed Laplacian Filter}

We next study the effect of edge perturbations on the signed Laplacian filter $g(\mathcal{G})=\mathbf{L}=\mathbf{D}-\mathbf{A}$. The output representation at each node $u$ aggregates the signed feature differences between $u$ and its neighbors:
$[\mathbf{LX}]_{u,:} = \sum_{ (u,v) \in \mathcal{E}^+ \cup \mathcal{E}^-} \mathbf{X}_{u,:} -s_{uv} \mathbf{X}_{v,:}$ 
where $s_{uv}=\mathbf{A}_{uv}\in\{+1,-1\}$ denotes the sign of edge $(u,v)$.

Similar to the signed adjacency filter, we analyze the effect of local edge perturbations by decomposing the signed Laplacian into individual edge contributions.  For a signed edge $e=(u,v,s_e)$ with $s_e\in\{+1,-1\}$, we define its signed incidence vector by $\mathbf{b}_e=\mathbf{e}_u-s_e\mathbf{e}_v\in\mathbb{R}^n$. The signed Laplacian can then be decomposed as
$\mathbf{L} =\sum_{e\in \mathcal{E}^+ \cup \mathcal{E}^-} \mathbf{b}_e\mathbf{b}_e^\top= \sum_{e\in \mathcal{E}^+ \cup \mathcal{E}^-}\mathbf{L}_e$, where we use $\mathbf{L}_e=\mathbf{b}_e\mathbf{b}_e^\top$ to denote the contribution of a single signed edge to the Laplacian.

To characterize edge perturbations under this decomposition, we decompose each perturbation into elementary signed-edge changes. An edge addition or deletion corresponds to one elementary change, whereas a sign flip corresponds to two: removal of the original signed edge and addition of the edge with the opposite sign.
Let $\mathcal{P}_L$ denote the resulting set of elementary changes. Each $e\in\mathcal{P}_L$ is associated with a coefficient $\alpha_e\in\{+1,-1\}$, where $\alpha_e=+1$ indicates that $\mathbf{L}_e$ is added and $\alpha_e=-1$ indicates that it is removed. The resulting filter perturbation can therefore be written as
$ \mathbf{E}_g
    =
    \mathbf{L}_p-\mathbf{L}
    =
    \sum_{e\in\mathcal{P}_L}
    \alpha_e\mathbf{L}_e.$ 
Substituting this decomposition into Eq.~\eqref{eq:signed_filter_energy}, the expected representation perturbation becomes
\begin{equation}
\begin{aligned}
    \mathbb{E}_{\mathbf{X}\sim\mathcal{D}}
    \left[
    \|\mathbf{E}_g\mathbf{X}\|_F^2
    \right]
    &=
    \left\langle
    \mathbf{K},
    \mathbf{E}_g^\top\mathbf{E}_g
    \right\rangle  
    =
    \sum_{e,e'\in\mathcal{P}_L}
    \alpha_e\alpha_{e'}
    \left\langle
    \mathbf{K},
    \mathbf{L}_e\mathbf{L}_{e'}
    \right\rangle .
\end{aligned}
\label{eq:proof_lap_expand_revised}
\end{equation}
This expansion again contains two types of terms: self terms with $e=e'$ and cross
terms with $e\neq e'$. 

For self terms, since $\|\mathbf{b}_e\|_2^2=1+s_e^2=2$, we have
$  \mathbf{L}_e^2 = \mathbf{b}_e
    (\mathbf{b}_e^\top\mathbf{b}_e)
    \mathbf{b}_e^\top  = 2\mathbf{L}_e .$ For each elementary signed-edge change $e=(u,v,s_e)$,  we have
$\left\langle \mathbf{L}_e,\mathbf{X}\mathbf{X}^\top \right\rangle =  \|\mathbf{X}^\top\mathbf{b}_e\|_2^2  = \|\mathbf{X}_{u,:}-s_e\mathbf{X}_{v,:}\|_2^2$. We therefore define the expected \emph{signed disagreement} associated with $e$ as
\begin{equation}
\label{eq:elemmetary_self}
    R_{s_e}(u,v)
    = \left\langle \mathbf{L}_e, \mathbf{K}\right\rangle
    = \mathbb{E}_{\mathbf{X}\sim\mathcal{D}} \left\langle \mathbf{L}_e,\mathbf{X}\mathbf{X}^\top \right\rangle= 
    \mathbb{E}_{\mathbf{X}\sim\mathcal{D}}
    \left[
    \|\mathbf{X}_{u,:}-s_e\mathbf{X}_{v,:}\|_2^2
    \right].
\end{equation}
It follows that $\left\langle
    \mathbf{K},
    \mathbf{L}_e^2
    \right\rangle
    = 2  \left\langle
    \mathbf{K},
    \mathbf{L}_e
    \right\rangle
    =  2R_{s_e}(u,v).$ Here, $R_{s_e}(u,v)$ quantifies the individual contribution of each elementary signed-edge change to the expected representation perturbation, where a larger value of $R_{s_e}(u,v)$ indicates greater disagreement between the node features and the signed relation represented by $e$. Since $\alpha_e^2=1$, the total self-contribution is
$ 2\sum_{e=(u,v,s_e)\in\mathcal{P}_L} R_{s_e}(u,v)$.

For two distinct elementary changes $e$ and $e'$, the product
$\mathbf{L}_e\mathbf{L}_{e'}$ depends on whether the two signed edges share an
endpoint. If
$e=(u,v,s_e)$ and $e'=(a,b,s_{e'})$ have no common endpoint, then
$\mathbf{b}_e^\top\mathbf{b}_{e'}=0$ by orthogonality of the standard basis
vectors. Hence $\mathbf{L}_e\mathbf{L}_{e'}=\mathbf{0}$, and the corresponding
cross term vanishes. It remains to analyze two elementary changes  that share exactly one endpoint $u$: $ e=(u,v,s_e), e'=(u,v',s_{e'}), v\neq v'$. Since
$\mathbf{b}_e^\top\mathbf{b}_{e'}=1$, we have
$\mathbf{L}_e\mathbf{L}_{e'}  = \mathbf{b}_e\mathbf{b}_{e'}^\top$. 
Therefore, $
\left\langle
\mathbf{K},
\mathbf{L}_e\mathbf{L}_{e'}
\right\rangle
=
\mathbb{E}
\left[
\left\langle
\mathbf{X}_{u,:}-s_e\mathbf{X}_{v,:},
\mathbf{X}_{u,:}-s_{e'}\mathbf{X}_{v',:}
\right\rangle
\right].
\label{eq:lap_cross_inner}
$ To rewrite this interaction in terms of signed disagreements, define the
\emph{cross signed disagreement} as
\begin{equation}
    R_{s_e,s_{e'}}(v,v')
    =
    \mathbb{E}_{\mathbf{X}\sim\mathcal{D}}
    \left[
    \|s_e\mathbf{X}_{v,:}-s_{e'}\mathbf{X}_{v',:}\|_2^2
    \right].
\end{equation}
By the polarization identity, we can get
\begin{equation}
\begin{aligned}
&
R_{s_e}(u,v)
+
R_{s_{e'}}(u,v')
-
R_{s_e,s_{e'}}(v,v') \\
&\qquad =
2\,
\mathbb{E}_{\mathbf{X}\sim\mathcal{D}}
\left[
\left\langle
\mathbf{X}_{u,:}-s_e\mathbf{X}_{v,:},
\mathbf{X}_{u,:}-s_{e'}\mathbf{X}_{v',:}
\right\rangle
\right].
\end{aligned}
\label{eq:proof_lap_cross_identity_revised}
\end{equation}
Thus, the two ordered terms $(e,e')$ and $(e',e)$ in
Eq. \eqref{eq:proof_lap_expand_revised} together contribute
$ \alpha_e\alpha_{e'}
    \Big[
    R_{s_e}(u,v)
    +
    R_{s_{e'}}(u,v')
    -
    R_{s_e,s_{e'}}(v,v')
    \Big]. $ Combining the self and cross terms yields the following decomposition.

\paragraph{Proposition 2}
\textit{For the signed Laplacian filter $g(\mathbf{\mathcal{G}})=\mathbf{L}$, the expected
embedding perturbation satisfies}
\begin{equation}
\begin{aligned}
    &\mathbb{E}_{\mathbf{X}\sim\mathcal{D}}
    \left[
    \|\mathbf{E}_g\mathbf{X}\|_F^2
    \right]
    =
    2
    \sum_{e=(u,v,s_e)\in\mathcal{P}_L}
    R_{s_e}(u,v)
    + \\
    &\sum_{\substack{
    \{e,e'\}\subseteq\mathcal{P}_L\\
    e\cap e'\neq\emptyset
    }}
    \alpha_e\alpha_{e'}
    \Big[
    R_{s_e}(u,v)
    +
    R_{s_{e'}}(u,v')
    -
    R_{s_e,s_{e'}}(v,v')
    \Big].
\end{aligned}
\label{eq:prop2}
\end{equation}

The first term is the self-contribution of each elementary
signed-edge change in Eq.\ref{eq:elemmetary_self}. It is large when the changed signed relation is incompatible
with the feature distribution.
For example, between positively correlated nodes, removing a positive edge, adding a negative edge, or flipping a positive edge to negative can involve elementary changes with large signed disagreement.
The second term captures interactions
between elementary changes that share a node. 
The bracket measures how the two signed relations jointly align with the features, while
$\alpha_e\alpha_{e'}$ determines whether the two add/remove directions reinforce
or cancel each other. 







\subsection{Experimental Setup}

\subsubsection{Datasets.}
We evaluate on synthetic and real-world signed graphs. Synthetic graphs are generated
from a two-community signed stochastic block model
(SSBM)~\cite{cucuringu2019sponge} with $n=50$ and sign-consistency
$c=0.8$: conditional on an edge existence, its sign is positive for nodes in the same community and negative for nodes in different communities with probability $c$, and is reversed with probability $1-c$.  Node features are sampled from community-dependent Gaussian distributions. For real-world dataset, we use the signed networks from Highland Tribes and Bitcoin \cite{bonchi2019discovering}. 



\subsubsection{Theory-guided perturbation and baselines.}
Guided by Eq~\eqref{eq:prop1} and Eq~\eqref{eq:prop2}, we design the signed-edge
perturbation method that maximizes the expected filter-induced embedding change.
We use greedy selection on small graphs~\cite{jiang2015stochastic} and a
projected gradient-based relaxation on large graphs~\cite{rosen1960gradient,
xu2019topology}. We compare against the following baselines: 1) Random, which samples $m$ edge perturbations
uniformly \cite{zhang2025stability}; 2) Degree, which targets high-degree endpoints without using the feature
covariance $\mathbf{K}$ \cite{tao2021single}; 3) Frustration, which increases sign-feature
inconsistency and targets structural imbalance rather than the filter objective
~\cite{cartwright1956structural,
bonchi2019discovering}. Unless otherwise specified, the perturbation budget is
$5\%$ of the original edge count.

\subsubsection{Evaluation metrics.}
Following~\cite{zhang2025stability}, we assess attack effectiveness using two task-agnostic metrics alongside one task-dependent metric. Embedding drift $|\Delta\mathrm{ED}|= $  $\|\mathbf{Z}_p-\mathbf{Z}\|_F/\|\mathbf{Z}\|_F$ measures the
relative change between signed filter embeddings on $\mathcal{G}$ and $\mathcal{G}_{p}$ calculated as 
Eqs.~\eqref{eq:prop1} and Eqs.\eqref{eq:prop2} in Section \ref{sec:theory}.
\emph{Polarity change} $|\Delta P|$= $|\beta(\mathcal{G})-\beta(\mathcal{G}_p)|$ uses the frustration-based
balance score $\beta(\mathcal{G})=1-f(\mathcal{G})/|E|$~\cite{bonchi2019discovering}, where $f(\mathcal{G})$ is
the minimum frustrated-edge count under a binary partition. For the \emph{task-dependent} setting, signed link prediction is a standard
benchmark~\cite{derr2018signed,huang2019sigat}. Using the chosen signed filter
(adjacency or Laplacian) as the encoder, we split the observed edges of $\mathcal{G}$ into
80\% training and 20\% test sets. The predictor is trained only on the clean training edges
only and then frozen. We evaluate AUC on the same held-out test edges under
$\mathcal{G}$ and $\mathcal{G}_p$ and report the resulting AUC drop $|\Delta\mathrm{AUC}|$.



\subsection{Experimental results}
\begin{table*}[t]
\centering
\caption{Synthetic and real-data results reported as the absolute change
(mean $\pm$ standard deviation).}
\label{tab:synthetic_real_data}
\resizebox{0.9\textwidth}{!}{
\begin{tabular}{ll ccc ccc}
\toprule
& & \multicolumn{3}{c}{Signed Adjacency} & \multicolumn{3}{c}{Signed Laplacian} \\
\cmidrule(lr){3-5} \cmidrule(lr){6-8}
Dataset & Method
& $|\Delta\mathrm{ED}|$ & $|\Delta P|$ & $|\Delta$AUC$|$
& $|\Delta\mathrm{ED}|$ & $|\Delta P|$ & $|\Delta$AUC$|$ \\
\midrule
\multirow{4}{*}{Synthetic}
& Random      & 0.195 $\pm$ 0.025 & 0.006 $\pm$ 0.008 & 0.008 $\pm$ 0.026
              & 0.598 $\pm$ 0.164 & 0.006 $\pm$ 0.008 & 0.031 $\pm$ 0.026 \\
& Degree      & 0.134 $\pm$ 0.058 & 0.001 $\pm$ 0.009 & 0.009 $\pm$ 0.049
              & 0.202 $\pm$ 0.116 & 0.001 $\pm$ 0.009 & 0.026 $\pm$ 0.024 \\
& Frustration & 0.029 $\pm$ 0.031 & 0.002 $\pm$ 0.003 & 0.002 $\pm$ 0.004
              & 0.120 $\pm$ 0.118 & 0.002 $\pm$ 0.003 & 0.006 $\pm$ 0.018 \\
& Ours      & \textbf{0.357 $\pm$ 0.068} & \textbf{0.010 $\pm$ 0.020} & \textbf{0.032 $\pm$ 0.047}
              & \textbf{1.400 $\pm$ 1.034} & \textbf{0.008 $\pm$ 0.031} & \textbf{0.078 $\pm$ 0.086} \\
\midrule
\multirow{4}{*}{Highland tribes}
& Random      & 0.221 $\pm$ 0.068 & 0.018 $\pm$ 0.014 & 0.018 $\pm$ 0.074
              & 0.309 $\pm$ 0.087 & 0.018 $\pm$ 0.014 & 0.013 $\pm$ 0.046 \\
& Degree      & 0.098 $\pm$ 0.024 & 0.002 $\pm$ 0.000 & 0.006 $\pm$ 0.022
              & 0.138 $\pm$ 0.033 & 0.002 $\pm$ 0.000 & 0.001 $\pm$ 0.025 \\
& Frustration & 0.092 $\pm$ 0.022 & 0.008 $\pm$ 0.008 & 0.000 $\pm$ 0.032
              & 0.237 $\pm$ 0.104 & 0.008 $\pm$ 0.008 & 0.001 $\pm$ 0.016 \\
& Ours      & \textbf{0.330 $\pm$ 0.064} & \textbf{0.038 $\pm$ 0.017} & \textbf{0.032 $\pm$ 0.068}
              & \textbf{0.516 $\pm$ 0.155} & \textbf{0.037 $\pm$ 0.017} & \textbf{0.061 $\pm$ 0.102} \\
\midrule
\multirow{4}{*}{Bitcoin}
& Random      & 0.194 $\pm$ 0.012 & \textbf{0.022 $\pm$ 0.001} & 0.007 $\pm$ 0.005
              & 0.039 $\pm$ 0.004 & 0.022 $\pm$ 0.001 & 0.004 $\pm$ 0.004 \\
& Degree      & 0.061 $\pm$ 0.143 & 0.003 $\pm$ 0.001 & 0.008 $\pm$ 0.040
              & 0.063 $\pm$ 0.044 & 0.003 $\pm$ 0.001 & 0.002 $\pm$ 0.004 \\
& Frustration & 0.103 $\pm$ 0.052 & 0.020 $\pm$ 0.001 & 0.011 $\pm$ 0.004
              & 0.120 $\pm$ 0.037 & 0.018 $\pm$ 0.002 & 0.009 $\pm$ 0.002 \\
& Ours      & \textbf{0.303 $\pm$ 0.061} & 0.021 $\pm$ 0.001 & \textbf{0.013 $\pm$ 0.007}
              & \textbf{0.201 $\pm$ 0.024} & \textbf{0.023 $\pm$ 0.001} & \textbf{0.010 $\pm$ 0.006} \\
\bottomrule
\end{tabular}
}
\vspace{-0.2cm}
\end{table*}

\subsubsection{Perturbation results}

Table~\ref{tab:synthetic_real_data} shows that the theory-guided perturbation
consistently induces the largest embedding drift under both filters and achieves
the strongest AUC degradation across all datasets. The polarity metric is more
mixed, especially on Bitcoin, indicating that global polarity change does not
fully explain representation instability. Overall, the results support the
proposed distribution-aware criterion, which identifies more disruptive
perturbations than baselines.

\subsubsection{Visualization of the perturbed edges}
\begin{figure}[t]
    \includegraphics[width=1.1\linewidth]{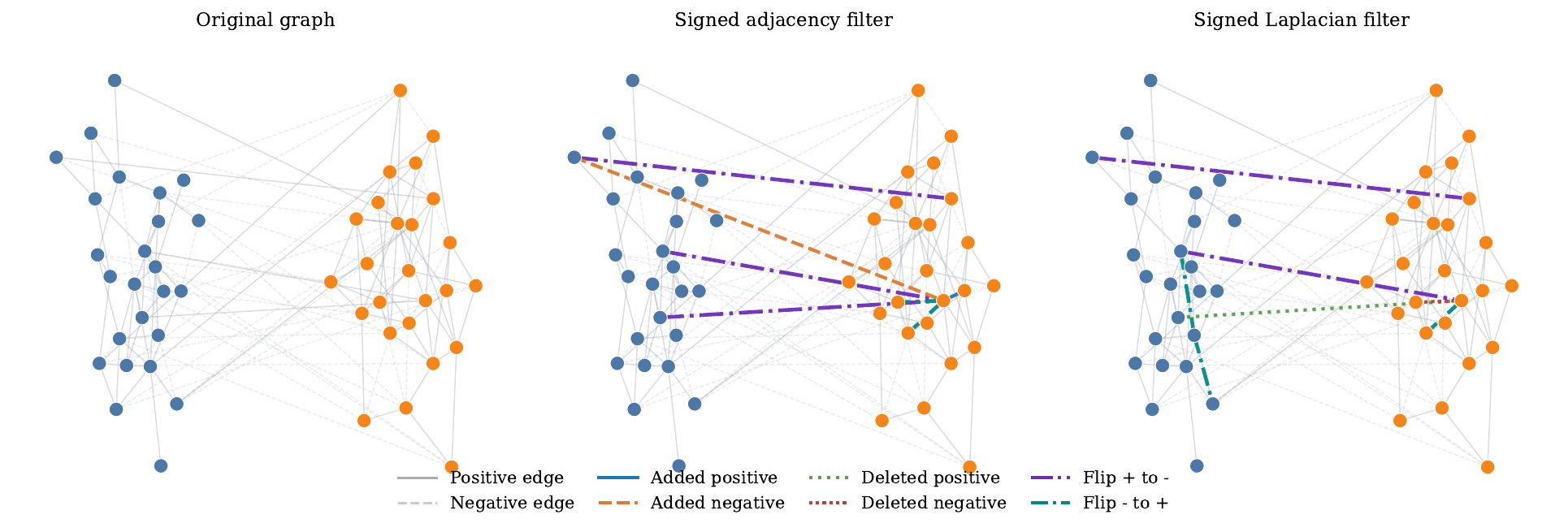}
    \caption{\small Visualization of perturbations selected by our proposed theory-guided attack on a synthetic signed graph.}
    \label{fig:visual_results}
    \vspace{-0.5cm}
\end{figure}

Figure~\ref{fig:visual_results} shows representative perturbations selected by
our theory-guided attack. The signed adjacency attack mainly targets
inter-community edges through sign flips, directly changing how
signals propagate between two polarized groups. Interestingly, these perturbations also induce larger changes in structural polarity: for both the signed adjacency filter and the signed Laplacian filter, the selected perturbations tend to promote positive relations within the same community and negative relations between different communities. Thus, the attack reinforces signed assortativity and increases polarization, which in turn produces substantial drift in the embeddings.

\begin{figure}[t]
\centering
\begin{subfigure}{0.48\linewidth}
    \centering
    \includegraphics[width=\linewidth]{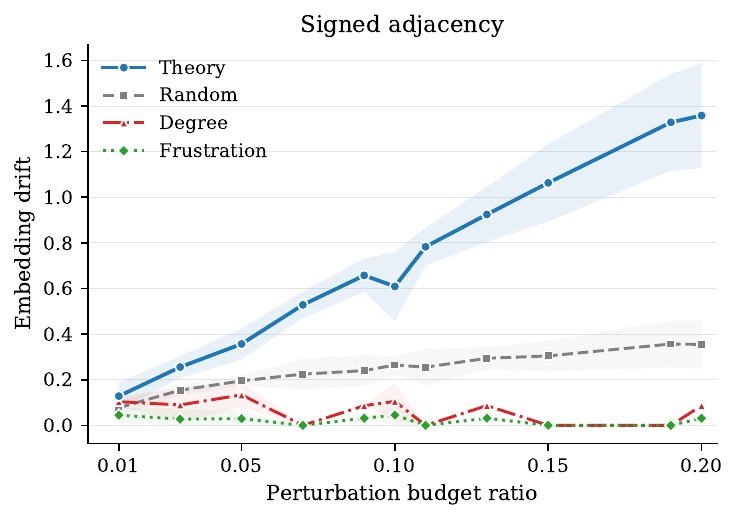}
    \caption{Signed adjacency: $|\Delta\mathrm{ED}|$}
    \label{fig:cloister_adj_drift}
\end{subfigure}
\begin{subfigure}{0.48\linewidth}
    \centering
    \includegraphics[width=\linewidth]{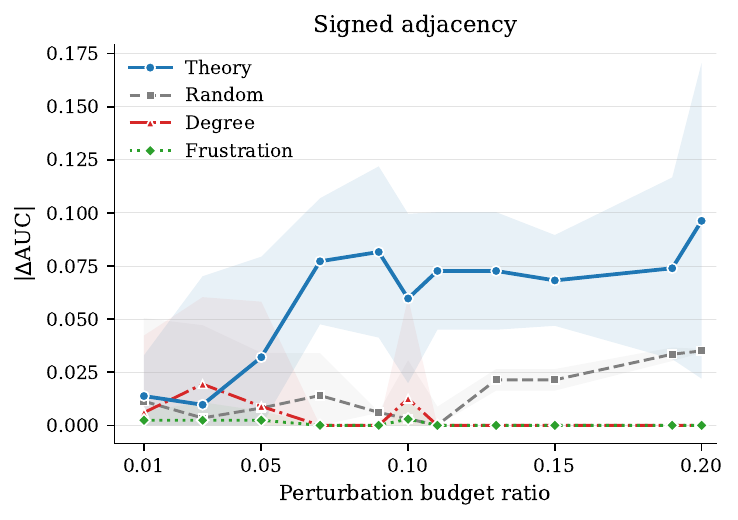}
    \caption{Signed adjacency: $|\Delta\mathrm{AUC}|$}
    \label{fig:cloister_adj_auc}
\end{subfigure}
\begin{subfigure}{0.48\linewidth}
    \centering
    \includegraphics[width=\linewidth]{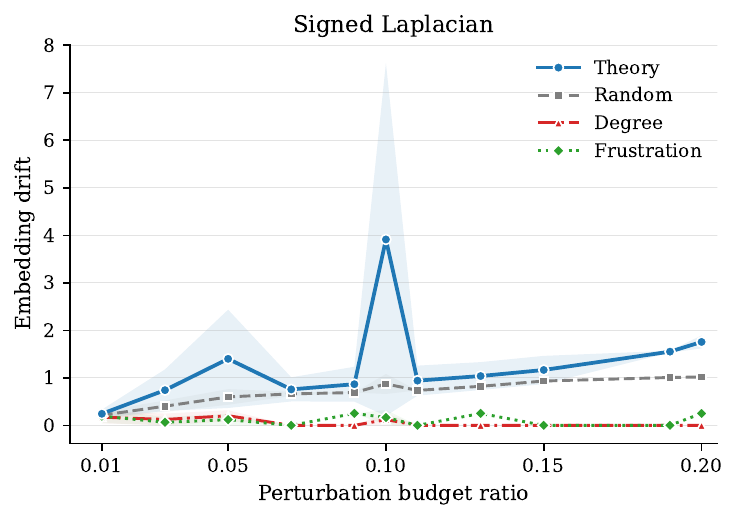}
    \caption{Signed Laplacian: $|\Delta\mathrm{ED}|$}
    \label{fig:cloister_lap_drift}
\end{subfigure}
\begin{subfigure}{0.48\linewidth}
    \centering
    \includegraphics[width=\linewidth]{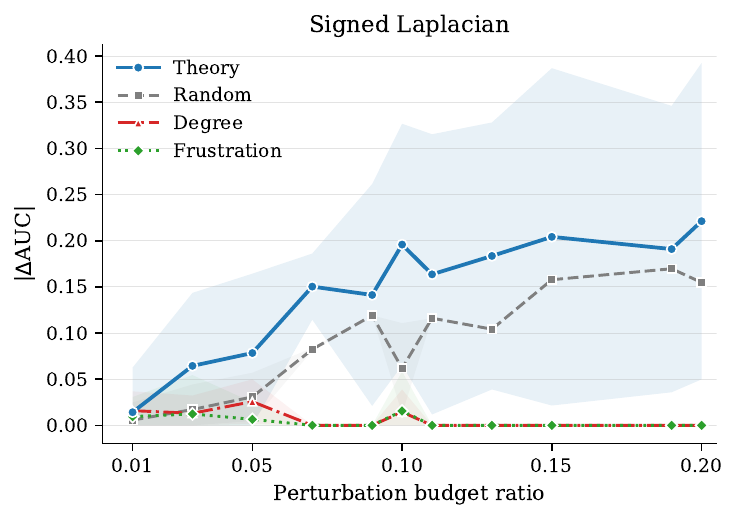}
    \caption{Signed Laplacian: $|\Delta\mathrm{AUC}|$}
    \label{fig:cloister_lap_auc}
\end{subfigure}

\caption{\small Budget trends for signed adjacency and signed laplacian filters. The $x$-axis is the perturbation budget ratio. The $y$-axis reports $|\Delta\mathrm{ED}|$ and $|\Delta\mathrm{AUC}|$.}
\label{fig:budget_trend}
\vspace{-0.5cm}
\end{figure}

\subsubsection{Budget sensitivity.}
Figure~\ref{fig:budget_trend} shows the effect of increasing the perturbation
budget. Across both signed adjacency and signed Laplacian settings, the proposed
theory-guided perturbation consistently achieves the largest embedding drift and
AUC degradation. Moreover, the gap between our method (Theory) and the baselines generally
grows with the budget ratio, indicating that the perturbation objective
continues to identify highly influential signed edges as additional perturbations
are allowed. 

\section{Conclusion}
In this paper, we studied the stability of signed graph filters under signed structural perturbations. We derived perturbation decompositions for signed adjacency and signed Laplacian filters, showing that their representation sensitivity depends jointly on edge changes, polarity, and node-signal statistics. Based on this analysis, we proposed a theory-guided perturbation strategy that identifies influential signed-edge modifications, including sign flips. Experiments on synthetic and real signed graphs show that the proposed method induces larger embedding drift and stronger downstream degradation than random, degree-based, and frustration-based baselines. These results highlight the importance of polarity-aware and distribution-aware robustness analysis for signed graph learning. Future work will extend the analysis to deeper signed GNNs and develop defenses against signed structural attacks.


\section{GenAI Usage Disclosure}
Generative AI tools were used solely for language editing and polishing of author-written text, including grammar, clarity, and wording improvements. All suggested edits were reviewed and revised by the authors. Generative AI was not used to develop the research ideas, methodology, code, data, experiments, results, or scientific conclusions of this work.

\bibliographystyle{ACM-Reference-Format}
\balance
\bibliography{cited}

\end{document}